\documentclass{ieeeaccess}
\usepackage{cite}
\usepackage{amsmath,amssymb,amsfonts}
\usepackage{bbm}
\usepackage{algorithmic}
\usepackage{graphicx}
\usepackage{textcomp}
\usepackage{multirow}
\def\BibTeX{{\rm B\kern-.05em{\sc i\kern-.025em b}\kern-.08em
    T\kern-.1667em\lower.7ex\hbox{E}\kern-.125emX}}
\begin{document}
\history{Date of publication xxxx 00, 0000, date of current version xxxx 00, 0000.}
\doi{10.1109/ACCESS.2017.DOI}

\title{Bridging the gap between Human Action Recognition and Online Action Detection}

\author{\uppercase{Alban Main de Boissiere}\authorrefmark{1}, 
\uppercase{Rita Noumeir}.\authorrefmark{1}}

\address[1]{Laboratoire de Traitement de l’Information en Santé, École de Technologie Supérieure, Montreal, QC H3C 1K3 (email: alban.main-de-boissiere.1@ens.etsmtl.ca, rita.noumeir@etsmtl.ca)}
\tfootnote{This work was supported by the Natural Sciences and Engineering Research Council of Canada, Prompt Québec, and Aerosystems International Inc.}

\markboth
{Author \headeretal: Preparation of Papers for IEEE TRANSACTIONS and JOURNALS}
{Author \headeretal: Preparation of Papers for IEEE TRANSACTIONS and JOURNALS}

\corresp{Corresponding author: Alban Main de Boissiere (e-mail: alban.main-de-boissiere.1@ens.etsmtl.ca).}

\begin{abstract}
Action recognition, early prediction, and online action detection are complementary disciplines that are often studied independently. Most online action detection networks use a pre-trained feature extractor, which might not be optimal for its new task. We address the task-specific feature extraction with a teacher-student framework between the aforementioned disciplines, and a novel training strategy. Our network, Online Knowledge Distillation Action Detection network (OKDAD), embeds online early prediction and online temporal segment proposal subnetworks in parallel. Low interclass and high intraclass similarity are encouraged during teacher training. Knowledge distillation to the OKDAD network is ensured via layer reuse and cosine similarity between teacher-student feature vectors. Layer reuse and similarity learning significantly improve our baseline which uses a generic feature extractor. We evaluate our framework on infrared videos from two popular datasets, NTU RGB+D (action recognition, early prediction) and PKU MMD (action detection). Unlike previous attempts on those datasets, our student networks perform without any knowledge of the future. Even with this added difficulty, we achieve state-of-the-art results on both datasets. Moreover, our networks use infrared from RGB-D cameras, which we are the first to use for online action detection, to our knowledge.
\end{abstract}


\titlepgskip=-15pt

\maketitle

\section{Introduction}
\label{sec:introduction}
\PARstart{C}{omputer} vision branches out to many subfields that are often studied independently. In the video understanding domain, action recognition aims at classifying entire segmented sequences. On the other end, action detection embodies the ability to detect and classify multiple activities inside a longer, unsegmented sequence. 

The emergence of consumer-grade depth cameras (RGB+D) \cite{keselman2017intel}, \cite{zhang2012microsoft} has sparked a new research dynamic in action understanding. Real-time pose estimation algorithms \cite{shotton2011real} are employed to extract 3D skeleton data from the depth stream. Additionally, RGB and infrared streams are also available. Except for the infrared, the various streams have been widely studied \cite{wang2018rgb}. In essence, the infrared and RGB representations are similar, with an advantage for the former. Infrared videos are represented on a gray scale and are noisy, which logically encourages the use of RGB data. But infrared is less affected by illumination conditions and is usable in the dark, an important property for security and night-vision applications.

\begin{figure*}[t]
  \centering
  \includegraphics[width=0.7\textwidth]{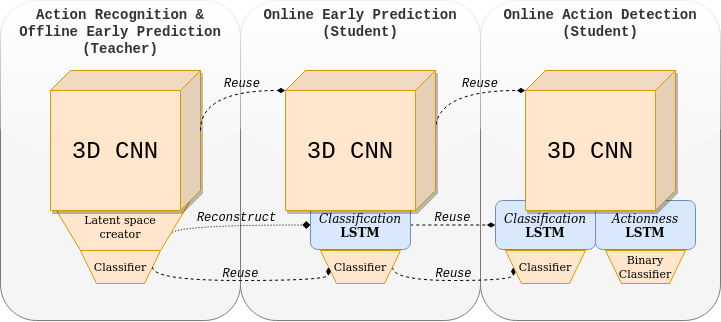}
  \caption{Action recognition to online action detection framework. A teacher is first trained. Its 3D CNN and classifier are reused by the students. Teacher feature vector reconstruction is encouraged via knowledge distillation. The online action detection student (OKDAD) embeds an additional temporal proposal module.}
  \label{fig:global_framework}
\end{figure*}

Action recognition is usually conducted by analyzing a sequence in its entirety before emitting a prediction. Early efforts leveraged recurrent neural networks (RNN) to study skeleton data as time series \cite{liu2016spatio}, \cite{shahroudy2016ntu}, \cite{zhang2017view}. This then shifted toward 2D convolutional neural networks (CNN) \cite{ke2017new}, \cite{kim2017interpretable}, \cite{zhang2019view} and graph convolutional networks (GCN) \cite{yan2018spatial}. But temporal normalization is always performed through sampling \cite{liu2016spatio}, \cite{shahroudy2016ntu}, \cite{zhang2017view}, image mapping and rescaling \cite{ke2017new}, \cite{kim2017interpretable}, \cite{zhang2019view} or global pooling \cite{yan2018spatial}. Temporal normalization is referred to as a step that normalizes the duration of a sequence. In other words, the duration of the sequence is known before it is studied. As such, most action recognition methods are considered offline, i.e. not applicable as the action happens or in real-time.

Early action prediction aims at recognizing a human activity before it is fully executed. The objective is therefore similar to the classification task of online action recognition. While some approaches are in line with the online paradigm \cite{ma2016learning}, \cite{sadegh2017encouraging}, recent attempts tackle the problem by dividing a sequence into $N$ shorter segments before evaluating a subset of these \cite{hu2018early}, \cite{pang2019dbdnet}, \cite{wang2019progressive}. In other words, the duration of the sequence is still known and used, which is considered offline. 

Online action detection evaluates raw, unsegmented sequences containing multiple actions \cite{de2016online}. The objective is to recognize an occurring activity and classify it frame by frame, or by groups of frames, as it happens. It differs from offline action detection where the sequence is studied in its entirety before temporal segments are proposed, then classified. Following the works of \cite{boissiere2020infrared}, we focus on infrared data from RGB-D cameras to further perpetuate their legitimacy as a stream candidate for action recognition and detection. Infrared from RGB-D cameras shows a strong potential for security and night-vision applications. In \cite{boissiere2020infrared}, infrared yields great results for action recognition. In this work, we go one step further and use infrared as a ready-to-embed network for real-time online action detection.

This work is motivated by the apparent natural progression from action recognition to online action detection that has not been previously exploited. A strong online early prediction network should translate well to online action detection. A network that is able to correctly classify an action as it happens should also be able to differentiate an action vs. no action in time. In other words, the online action detection problem can be broken down into two parallel online early prediction tasks. The first task is to classify video frames as containing an action or not. The second is to classify the possible ongoing action. In essence, online action detection can be expressed as binary online early prediction to detect actions and conventional online early prediction to classify them.  

As such, we focus our efforts on online early prediction which are most frequently evaluated on the NTU-RGB+D dataset \cite{shahroudy2016ntu} instead of common online action detection datasets \cite{de2016online}, \cite{idrees2017thumos}. Additionally, our work heavily borrows elements from \cite{boissiere2020infrared} which uses infrared from RGB-D cameras.

We propose a framework (Fig. \ref{fig:global_framework}) to link the above-mentioned research problems, using a 3D CNN and long short-term memory (LSTM) RNNs as building blocks. We deploy an offline network to tackle both action recognition and early prediction (teacher). We use it to distill knowledge to an early prediction student network, which is online, likewise for an online action detection student. The classification LSTM is used to recognize an ongoing action for both students. The "actionness" LSTM, used only for online action detection, classifies a block of frames as containing an action or not. From here on out, we refer to actionness as the likelihood of a frame-block being an action. We train our network on infrared videos from RGB+D cameras.

Our main contributions are as follows: 1) A novel framework for both RGB-D human action recognition and offline early prediction with infrared videos; 2) Cosine similarity loss terms, which improve teacher accuracy; 3) A teacher-student knowledge distillation framework for offline to online early prediction based on cosine similarity; 4) An online action detection architecture which builds upon the online student network; 5) State-of-the-art results and extensive experiments on infrared videos from two popular benchmark datasets \cite{liu2017pku}, \cite{shahroudy2016ntu}. 

The project code will be publicly available upon acceptance of the paper. Video demonstrations and further illustrations will also be available on the project page.

\section{Related work}

\subsection{Human action recognition}

Pioneering approaches to video action recognition used handcrafted spatiotemporal features, such as scale-invariant feature transform, histogram of oriented gradients, and improved Dense Trajectories, which are still competitive \cite{wang2013action}.

Recent efforts shifted toward deep learning. In \cite{karpathy2014large}, different temporal fusing schemes are explored, with 2D CNNs as spatial feature extractors. In \cite{simonyan2014two}, a two-stream network models spatiotemporal features via RGB images and optical flow. Temporal dependencies may also be modeled via recurrent networks \cite{donahue2015long}, \cite{yue2015beyond}. A 2D CNN outputs a feature vector for each frame, or group of frames, which is then fed to an LSTM network. In \cite{donahue2015long}, the CNN is pre-trained and frozen during training, which might not be ideal as the CNN cannot learn in the context of the video. Another family of networks is 3D CNNs \cite{carreira2017quo}, \cite{tran2015learning}, \cite{tran2018closer}, \cite{xie2018rethinking}. Their major drawback is the number of trainable parameters. In \cite{tran2018closer}, an architecture called ResNet (2+1)D (R(2+1)D) uses skip connections as its 2D counterpart \cite{he2016deep}. This leads to fewer parameters to optimize while retaining state-of-the-art performances. Additionally, spatial and temporal convolutions are separated by nonlinear activation functions to allow for a more complex representation.

Skeleton data are powerful \cite{johansson1973visual}, but superiority against video data is unclear; rather, they seem to be complementary \cite{boissiere2020infrared}, \cite{wang2018rgb}. First modern deep learning attempts gravitated toward various forms of RNNs \cite{liu2016spatio}, \cite{shahroudy2016ntu}, \cite{zhang2017view}. Then skeleton to 2D image mapping with 2D CNNs yielded better results \cite{ke2017new}, \cite{kim2017interpretable}, \cite{zhang2019view}. Nowadays, graph convolutional networks are the state of the art \cite{yan2018spatial}, \cite{shi2019skeleton}.

\subsection{Early action prediction}

Early action prediction follows the paradigm of action recognition, but on partially observed sequences. The first attempts used handcrafted features in the form of representation of visual words \cite{kong2014discriminative}, hierarchical movemes \cite{lan2014hierarchical} and histogram of spatiotemporal features \cite{ryoo2011human}.

Deep learning approaches can be separated into two categories: online and offline. Online approaches do not use the duration of a sequence as part of a preprocessing step. In \cite{ma2016learning}, a CNN+LSTM network is used. A novel loss is introduced which encourages a monotonic ascending prediction score over time. In \cite{sadegh2017encouraging}, a CNN+LSTM network is used for very early prediction, but can at most study 50 frames. A similar architecture is used in \cite{kong2018action}, but is offline because of the bidirectional nature of the RNN. 

The offline methods either preprocess data in the temporal domain \cite{ke2019learning}, \cite{wang2019progressive}, or the use information from the future \cite{ke2019learning}, \cite{kong2018action}, \cite{pang2019dbdnet}. In \cite{ke2019learning}, partial and full sequences are confronted in an adversarial learning context on handcrafted skeleton images. In \cite{wang2019progressive}, a bidirectional CNN+LSTM teacher distills its information to a unidirectional student network. The architecture could be online, but a temporal normalization step is first performed. 

\subsection{Action detection}

Action detection analyzes raw sequences and outputs temporal segments containing activities with a class prediction. The online framework emits a prediction frame by frame, or by short blocks of frames, without future context. In an offline scenario, a sequence is studied in its entirety before outputting predictions. 

\subsubsection{Offline action detection}

Early efforts focused on handcrafted features and sliding windows of different sizes \cite{caba2016fast}, \cite{ni2016progressively}, \cite{wang2014action}, \cite{yuan2016temporal}. Deep learning attempts follow the framework of the R-CNN family for object detection in images \cite{girshick2014rich}, \cite{girshick2015fast}, \cite{ren2015faster}. The network outputs temporal proposals, ranks them, then classifies them. The architectures presented in \cite{chao2018rethinking}, \cite{dai2017temporal}, \cite{gao2017turn}, \cite{gao2017cascaded}, \cite{xu2017r} combine those tasks in an end-to-end fashion. In \cite{song2018spatio}, proposal precedes classification via spatiotemporal attention LSTMs on skeleton data. 

Closer to the online efforts, some architectures study sequences in a single streamflow. In \cite{sstad_buch_bmvc17} and \cite{buch2017sst}, a 3D CNN+RNN architecture studies videos by groups of $\delta$ frames. The network in \cite{sstad_buch_bmvc17} could be used in real-time, but is limited by a fixed maximum proposal size and demanding post-processing.

\subsubsection{Online action detection}

De Geestet \textit{et al.} outlined the challenges of online action detection \cite{de2016online} and later proposed a two-stream LSTM architecture \cite{de2018modeling}. The first stream interprets the input, the other the temporal dependencies between actions. In \cite{gao2017red}, a Reinforced Encoder-Decoder (RED) network uses a CNN feature extractor with an LSTM. The network is designed for anticipation, but can be used for online detection. In \cite{shou2018online}, the precise start time of an action is emphasized with adversarial networks. In \cite{xu2019temporal}, a temporal recurrent network (TRN) is introduced with a prediction module. Using a pre-trained feature extractor yields good results in \cite{sstad_buch_bmvc17}, \cite{gao2017red}, \cite{xu2019temporal}, but we believe fine-tuning a network in the context of its new task leads to improved performances, as demonstrated in \cite{luo2018graph}

\begin{figure}[t]
\centering
\includegraphics[width=0.50\textwidth]{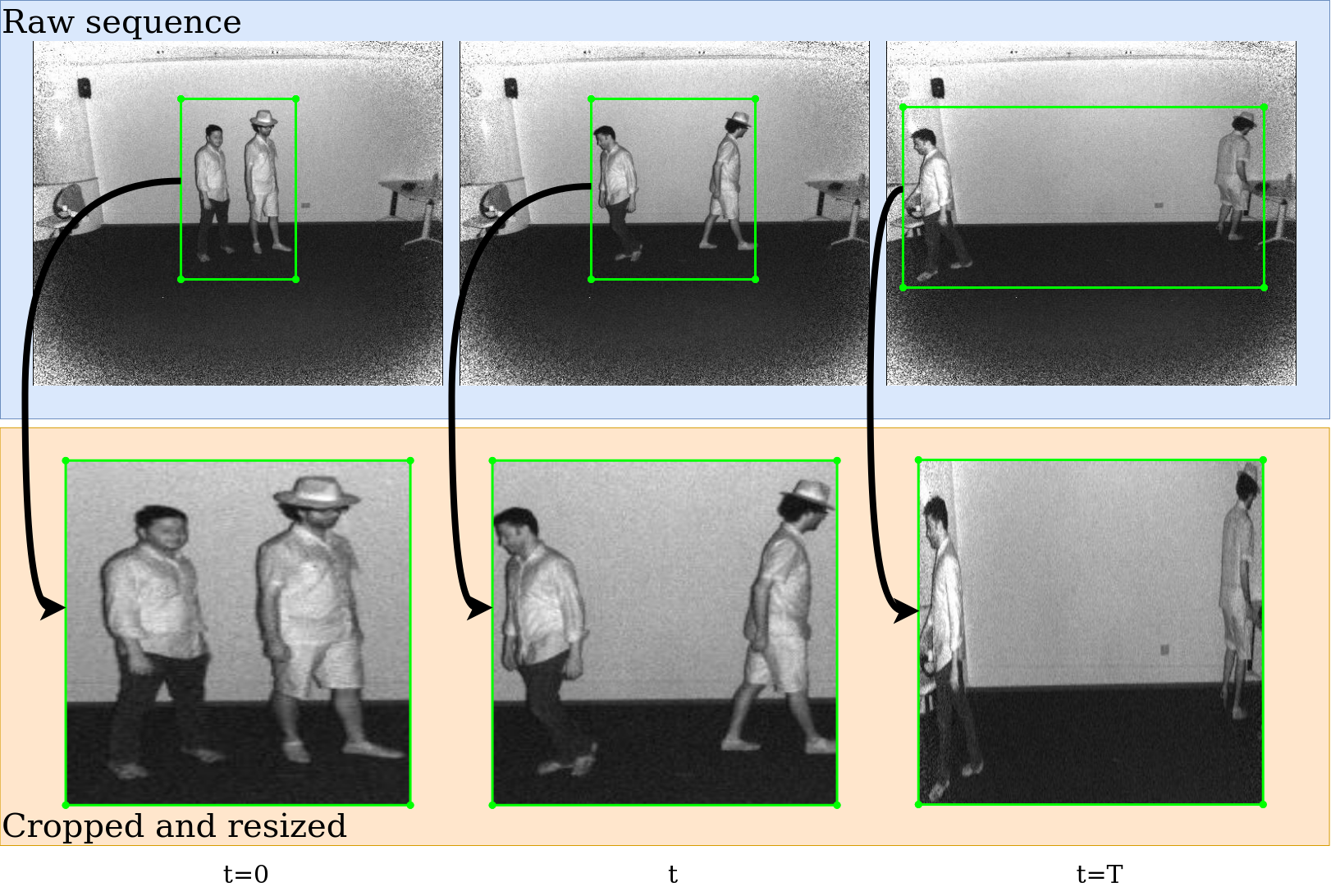}
\caption{Online preprocessing on infrared videos. A crop around the subjects is performed using 2D skeleton data.}
\label{fig:online_ir_preprocessing}
\end{figure}

\begin{figure*}[t]
    \centering
    \includegraphics[width=0.8\textwidth]{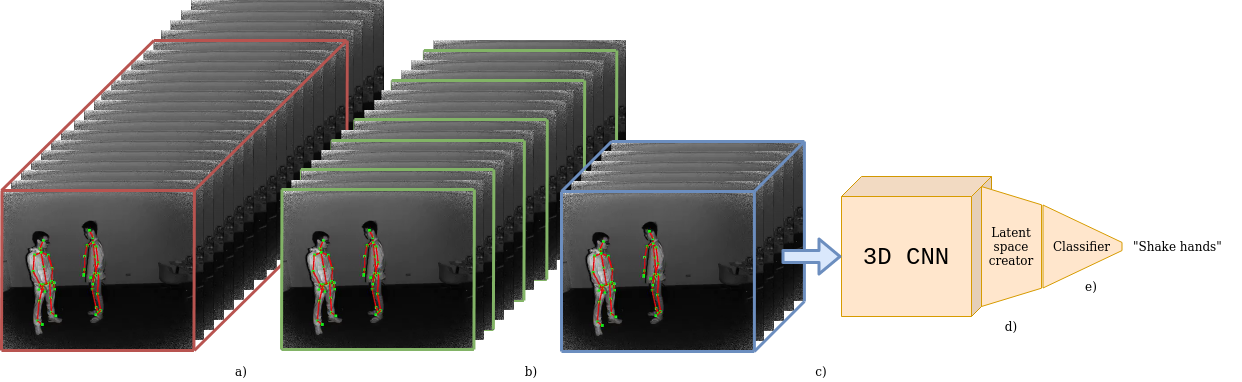}
    \caption{Teacher framework. a) A random observation ratio is used during training. b) $T^{off}$ frames are sampled from $T^{off}$ subwindows of even sizes. c) The normalized sequence is fed to a ResNet (2+1)D. d) The feature vector is normalized by a 1D batch normalization layer. e) The normalized teacher feature vector $x^{p}$ is used for prediction.}
    \label{fig:offline_action_prediction_framework}
\end{figure*}

\subsection{Knowledge Distillation}

Knowledge distillation regroups techniques that aim to transfer the knowledge of a large pre-trained network to a smaller one. The concept was introduced in \cite{bucilua2006model}. Popularized in \cite{hinton2015distilling}, "softmax temperature" is proposed. The student network learns from both the ground truth and smoothed soft labels from the teacher. Minimizing the mean square error (MSE) between student and teacher outputs is a possibility \cite{romero2014fitnets}. In \cite{yim2017gift}, knowledge distillation shows a faster learning time for the student network, which eventually outperforms the teacher. In \cite{wang2019progressive}, the loss function minimizes the maximum mean discrepancy (MMD) between teacher and student for early action prediction. Our approach borrows elements from both transfer learning and knowledge distillation. Also, we shift the focus from an offline teacher to online students. Because we do not only reuse a network but attempt to distill the teacher's representation, our framework also falls under the knowledge distillation paradigm. 

\section{Action recognition to online action detection framework}

We tackle multiple video understanding tasks, from human action recognition to online action detection, and propose a framework to link those together. An offline teacher is deployed for action recognition and early prediction. During training, intraclass cosine similarity and interclass cosine dissimilarity are encouraged. Knowledge distillation is employed to transfer the representation of the teacher to an online early prediction student network. Distillation is done via the reuse of layers and cosine similarity between teacher and student feature vectors at different time points. The student network is then transposed to an online action detection task. We introduce a sigmoid-weighted temporal loss to train the online networks.

\begin{figure*}[t]
\centering
\includegraphics[width=0.8\textwidth]{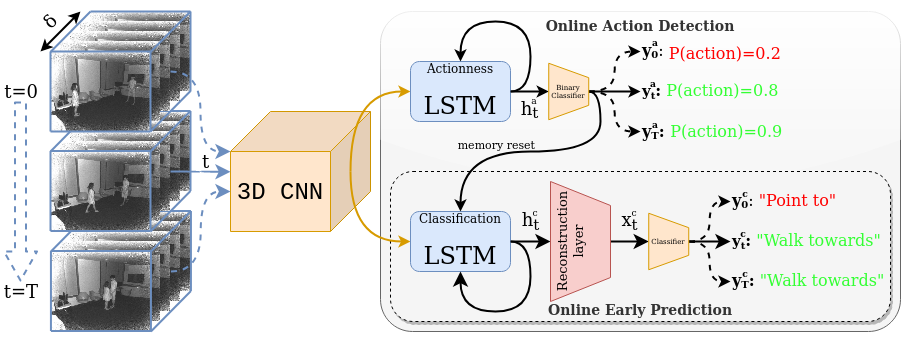}
\caption{Student networks. A sequence is broken down into chunks of $\delta$ frames fed to a 3D CNN. The actionness LSTM proposes temporal segments. The classification LSTM outputs a prediction when an action is being detected. The network can be used for online early action prediction with the classification LSTM only.}
\label{fig:online_action_prediction_detection}
\end{figure*}

\subsection{Preprocessing}

\subsubsection{Cropping strategy}

With online action detection as our final objective, we implement an online cropping strategy on the infrared frames, inspired by \cite{boissiere2020infrared}. This acts as a hard attention mechanism by focusing the action on the subjects performing it. It is used across the different networks. Also, the teacher and student networks will study sequences differently. The teacher requires a temporal normalization step, while the students do not, which leads to different sampling strategies.  

Because 3D CNNs embed many trainable parameters, the video frame resolution is heavily downscaled to offset the memory needs. This results in a non-negligible loss of information. For example, a small object might become indistinguishable. 

In a human action recognition context, the background provides little context regarding the activity performed. Thus, we use the 3D skeleton coordinates on the infrared frames to create a region of interest around the subject(s). As such, we extract the maximal and minimal pixel values across all joints at each time frame to capture the subject in a bounding box (Fig. \ref{fig:online_ir_preprocessing}). 

Because a frame is cropped before the resizing operation, the downscaling factor is less important, resulting in less information loss. However, because the bounding box is recalculated for every frame, it moves across time. This results in a fake camera movement and different scaling factors over time, as shown Fig. \ref{fig:online_ir_preprocessing}. Performances are affected because pixels lose their temporal coherence.

\subsubsection{Sampling strategies}

We define offline and online sampling strategies for the teacher and online networks respectively.

For action recognition, an action sequence is divided into $T^{off}$ subwindows of equal sizes, as done in \cite{boissiere2020infrared}. A random frame is sampled from each, creating a normalized sequence of length $T^{off}$ which is fed to the teacher. 

For the offline early prediction task, the total number of frames $N$ of a sequence is adjusted depending on the observation ratio $r$, i.e., the percentage of the sequence considered: $N_{partial}=\text{floor}(rN)$. For instance, for a sequence of $N=100$ total frames and an observation ratio of $r=80\%$, we will consider the $N_{partial}=80$ first frames. The same strategy employed for action recognition is then used with the adjusted number of frames. If $T^{off}$ is greater than $N_{partial}$, the first $N_{partial}$ frames are considered. The end of the resulting sequence is padded with black frames to reach a size $T^{off}$. For example, if $T^{off}$ is set to 15, and $N_{partial}=10$, then 5 black frames are added to get a normalized sequence of size $T^{off}$.

For the online networks (online early prediction and online action detection), the sequences are studied in their entirety. The frame rate is reduced by a factor $s$, meaning we keep one of every $s$ frames.

\subsection{Network architectures}

The building blocks for our architectures are a 3D CNN, LSTMs, and fully connected layers. They are reused across the different networks which improves both results and learning times.

\subsubsection{Offline teacher}

We use an 18-layer deep ResNet (2+1)D (R(2+1)D) as our backbone \cite{xie2018rethinking}. The network is pre-trained on Kinetics-400 \cite{carreira2017quo}. Our offline early prediction framework is summarized Fig. \ref{fig:offline_action_prediction_framework}. The R(2+1)D network outputs a feature vector of length 512 which we then normalize with a 1D batch normalization layer \cite{ioffe2015batch}. Batch normalization allows for all feature vectors to have roughly the same Euclidean norm. As such, minimizing intraclass cosine similarity should also reduce the MSE between feature vectors. We call $x^{p}$ the normalized teacher feature vector. A final fully connected layer, the classifier, is used to output a softmax-normalized class probability distribution.

\subsubsection{Online early prediction student}

The online early prediction student network reuses the same R(2+1)D as the teacher but adds a classification LSTM network on top (Fig. \ref{fig:online_action_prediction_detection}). For early prediction alone, it does not embed the actionness LSTM.  A sequence is divided into $T^{on}$ subsequences of $\delta$ frames, with $T^{on}=\text{ceil}(\frac{N}{\delta})$. For instance, with $\delta=5$, a sequence of $N=98$ frames, then $T^{on}=20$. Each subsequence is fed to the R(2+1)D network. At each time step $t \in \{1, .., T^{on}\}$, the computed feature vector is used as input for the LSTM. The classification LSTM hidden vector $h^c_t$ will then be fed to a "reconstruction" fully connected layer. From here on out, the superscript $c$ refers to the classification LSTM. We call the outputted vector $x^c_t$ at each time step $t$. During training, the goal will be to approximate $x^p_t$ with $x^c_t$. Here, $x^p_t$ is the feature vector outputted by the teacher with $N_{partial}=ts\delta$. 

\subsubsection{OKDAD student}

We introduce the Online Knowledge Distillation Action Detection (OKDAD) network (Fig. \ref{fig:online_action_prediction_detection}). It builds upon the previous student network with an additional actionness LSTM layer. The actionness module has two roles. Firstly, it proposes temporal segments as the sequence happens. Secondly, it resets the cell and hidden state vectors of the classification LSTM. Intuitively, it forces the classification LSTM to forget about the past once an action is over. As such, when a new action is discovered, the classification LSTM is reduced to an online early prediction task.

The hidden state vector of the actionness LSTM $h^a_t$ (the superscript $a$ refers to the actionness LSTM) is fed to a fully connected layer. Its output is followed by the sigmoid function, to predict the probability that the current frame block contains an action. With $y^a_t$ the actionness probability, the classification LSTM is updated as follows:
\begin{align} \label{eq:mem_reset}
    h^c_{t-1} &:= y^a_t h^c_{t-1} \\
    c^c_{t-1} &:= y^a_t c^c_{t-1}.
\end{align}

\subsection{Knowledge distillation}

\subsubsection{Teacher loss}

Recall that we aim to distill teacher knowledge to an online action detection network. We hypothesize that the reconstruction task of the student will be easier if the teacher's feature vectors have low intraclass and high interclass distance. We introduce loss terms based on cosine similarity, as we believe it is more appropriate than MSE for that task. In essence, it is more practical to "push apart" vectors of different classes with cosine similarity. Additionally, because the batch normalization layer implies that feature vectors have roughly the same Euclidean norm, encouraging cosine similarity for vectors of same classes should also reduce MSE, without explicitly penalizing it. Also, distance between vectors becomes harder to interpret in high feature dimensional spaces, 512 in our case. Therefore, we propose a novel loss function:

\begin{small}
\begin{align}
\label{eq:teacher_loss}
\text{loss} =& \text{r}^\gamma [\text{cross-entropy}
        + \text{similarity loss}
        + \text{dissimilarity loss}] \\
\text{loss} =& r^\gamma [\frac{1}{|B|} \sum_{i=1}^{|B|} H(\hat{y}_i, y^{p}_i) \\
        &+ \frac{\alpha}{N_=} 
        \sum_{i=1}^{|B|} (\sum_{j=i+1}^{|B|} - \mathbbm{1}_{\hat{y}_i=\hat{y}_j} \ln(\frac{\cos({\angle ({x^p_i, x^p_j}})) + 1 }{2}))  \nonumber\\
        &+ \frac{\beta}{N_{\neq}} 
        \sum_{i=1}^{|B|} (\sum_{j=i+1}^{|B|} - \mathbbm{1}_{\hat{y}_i\neq \hat{y}_j} \ln(1 - \frac{\cos({\angle ({x^p_i, x^p_j}})) + 1} {2}))].  \nonumber
\end{align}
\end{small}

$N_=$ and $N_{\neq}$ are respectively the number of pairs of a same class, and different classes for a given batch:

\begin{small}
\begin{align}
    N_{=} &= \sum_{i=1}^{|B|} (\sum_{j=i+1}^{|B|}  \mathbbm{1}_{\hat{y}_i=\hat{y}_j}) \\
    N_{\neq} &= \sum_{i=1}^{|B|} (\sum_{j=i+1}^{|B|} \mathbbm{1}_{\hat{y}_i\neq \hat{y}_j}).
\end{align}
\end{small}

And:

\begin{small}
\begin{equation}
    \mathbbm{1}_{condition} = \left\{
    \begin{array}{ll}
        1 & \mbox{if condition is true} \\
        0 & \mbox{otherwise.}
    \end{array}
\right.
\end{equation}
\end{small}

Also, $B$ is a batch of sequences, $\angle$ is the angle between two vectors $\hat{y}$ the ground truth, $y^{p}$ the teacher prediction, $x^p$ the normalized teacher feature vector, $H$ the cross-entropy loss, $r$ the observation ratio, $|B|$ the number of sequences in a batch, $\alpha$, $\beta$, and $\gamma$ three hyperparameters.

The loss consists of three terms. The cross-entropy loss, $H$, encourages correct predictions with a bias in favor of high confidence. The similarity loss term reduces cosine similarity between same-class vectors in a batch. The dissimilarity loss term decreases cosine dissimilarity between different-class vectors in a batch. Hyperparameters $\alpha$ and $\beta$ weigh the importance of similarity and distance loss terms respectively. Hyperparameter $\gamma$, taken from $]0, +\infty[$, modulates the importance of a smaller observation ratio. The smaller the value, the more penalty is applied.

\subsubsection{Online early prediction student loss}

We want our student network to mimic the teacher feature vector $x^{p}_t$ generated at different time steps $t \in \{1, .., T^{on}\}$. We propose a loss function that allows the network to learn autonomously while being guided by the teacher's correct predictions: 

\begin{small}
\begin{equation}
\begin{aligned} \label{eq:student_loss}
\text{loss} = \frac{1}{|B|} \sum_{i=1}^{|B|}& L_s(\hat{y}_i, y^c_i, x^{p}_i, x^{c}_i); \\
\text{loss} = \frac{1}{|B|} \sum_{i=1}^{|B|} &\left[\sum_{t=1}^{T^{on}_i} 
        \sigma_{i,t} [H(\hat{y}_{i,t}, y^c_{i,t}) \right. \\
         &\left.- \eta(1-\epsilon_{i,t}) \ln(\frac{\cos({\angle ({x_{i,t}^{p}, x_{i,t}^{c}}}))+1}{2})]\right].
\end{aligned}
\end{equation}
\end{small}

Again, $B$ is a set of training examples, i.e. a batch, and $i$ the index of given element. Here, $\epsilon_t$ is the teacher error, such that $\epsilon_t = 1 - P(y_t^{p} = \hat{y})$, where $P(y_t^{p} = \hat{y})$ is the softmax-generated probability from the teacher for the correct label. Weighted arithmetic average coefficients are sampled evenly from the sigmoid function in the $[-2, 2]$ range: $\sigma_t=\frac{2}{T^{on}} \sigma(4\frac{t-1}{T^{on}-1}-2)$ and $\sum_{t=1}^{T^{on}}\sigma_t=1$. The network is thus less penalized on early outputs. Notice that $T^{on}$ is variable. If $T_i^{on} = 1$, then we take $\sigma_{i,t}=1$. The hyperparameter $\eta$ weighs the contribution of the teacher.

The loss consists of two terms. The first one dynamically penalizes classification based on total sequence length. The second one guides the student's reconstruction vector toward the teacher's normalized feature vector. To further encourage similar representations, the teacher's classification layer is reused by the student, and frozen during training.

\subsubsection{OKDAD student loss}

During the training of the OKDAD network, two tasks are optimized. The first is the temporal segment propositions. This is ensured by minimizing the binary cross-entropy $H$ between predicted and true actionness at each frame-block. The second task is early prediction for each action in a studied sequence, with $A$ a set of actions and $|A|$ its cardinality. To do so, our sigmoid weighted temporal loss $L_s$ is reused (\ref{eq:student_loss}). We propose the following loss function:

\begin{equation}
\begin{aligned}
\text{loss} = \frac{1}{|B|} \sum_{i=1}^{|B|} &\left[ \frac{1}{T} \sum_{t=1}^T H(\hat{y}^a_{i,t}, y^a_{i,t}) \right . \\
&\left .+ \sum^{|A_i|}_{a=1} L_s(\hat{y}_{i,a}^c, y_{i,a}^c, x^{p}_{i,a}, x^{c}_{i,a})
\right]
\end{aligned}
\end{equation}

Intuitively, the OKDAD network learns to detect ongoing actions. When an action is detected, classifying it as it happens is equivalent to the online early prediction task. Recall the classification LSTM is only instantiated when an action is ongoing. In this regard, the second loss term only penalizes frame-blocks containing an action.

To summarize, the progression from action recognition to online action detection is as follows. First, an action detection dataset is reduced to an offline early prediction task which a teacher network learns to execute. The knowledge is distilled to an online action detection network, which we showed can be considered as an online early prediction task doubled with temporal action proposals.

\section{Experiments}
We test our method on infrared videos from RGB+D human action datasets: NTU RGB+D \cite{shahroudy2016ntu} and PKU-MMD \cite{liu2017pku}. To the best of our knowledge, they are the only RGB+D datasets including the infrared stream. NTU RGB+D is a human action recognition dataset on which we evaluate our offline teacher and the classification performances of our OKDAD architecture. PKU-MMD is a human action detection dataset, on which we evaluate both the temporal action proposal and classification tasks of OKDAD.

We focus our efforts on the online classification task. We believe a strong online early prediction network should naturally perform well for temporal action propositions, as it can be seen as a binary classification task.

\subsection{Implementation details}

\begin{table*}[t]
\begin{center}
\caption{Early prediction results on NTU-RGB+D (accuracy in \%). Bold values are the best.}
\label{table:early_pred_results}
\begin{tabular}{c|c|c|c|c|c|c|c|c|c|c|c}
Observation ratio & 10\% & 20\% & 30\% & 40\% & 50\% & 60\% & 70\% & 80\% & 90\% & 100\% & Avg. \\
\noalign{\smallskip}
\hline \hline
\noalign{\smallskip}
KNN \cite{hu2018early} & 7.4 & 9.5 & 12.2 & 16.0 & 20.8 & 25.9 & 30.8 & 34.4 & 36.1 & 37.0 & 23.01 \\
RankLSTM \cite{ma2016learning} & 11.5 & 16.4 & 25.6 & 37.7 & 47.9 & 55.9 & 60.9 & 64.4 & 66.0 & 65.9 & 45.22 \\
DeepSCN \cite{kong2017deep} & 16.8 & 21.4 & 30.5 & 39.9 & 48.7 & 54.6 & 58.1 & 60.1 & 60.0 & 58.6 & 44.87 \\
MSRNN \cite{hu2018early} & 15.1 & 20.3 & 29.5 & 41.3 & 51.6 & 59.1 & 63.9 & 67.3 & 68.8 & 69.2 & 48.61 \\
PTSL \cite{wang2019progressive} & 27.8 & \textbf{35.8} & 46.2 & 58.4 & 67.4 & 73.8 & 77.6 & 80.0 & 81.4 & 82.0 & 63.04 \\
DBNet \cite{pang2019dbdnet} & \textbf{27.9} & 33.3 & 47.2 & 56.9 & 68.5 & 74.5 & 78.5 & 80.5 & 81.6 & 81.5 & 63.04 \\
\noalign{\smallskip}
\hline
\noalign{\smallskip}
\textbf{Teacher} & 10.5 & 29.5 & \textbf{49.9} & \textbf{66.7} & \textbf{76.0} & \textbf{81.1} & \textbf{83.9} & \textbf{85.4} & \textbf{86.3} & 86.4 & \textbf{65.57} \\
\textbf{Student} & 19.3 & 28.1 & 38.6 & 55.5 & 67.9 & 75.5 & 80.6 & 83.7 & 85.5 & \textbf{86.6} & 62.13 \\ 
\noalign{\smallskip}
\end{tabular} 
\end{center}
\end{table*}
\setlength{\tabcolsep}{1.4pt}

An R(2+1)D network \cite{xie2018rethinking} is used across all networks. It outputs a 1D feature vector of length 512. For the student, the classification and actionness RNNs are single LSTMs with 2048 and 1024 features in the hidden state respectively. The reconstruction layer is a single linear layer with an input size of 2048 and an output size of 512.    

During teacher training, we sample the sequence observation ratio $r$ uniformly between $[r_{min}, 2]$. If the ratio is greater than one, it is set to one. This favors training on entire sequences. We use $r_{min}=0.025$ and use at least one frame. For best results, we use $\alpha=1$, $\beta=0.5$ and $\gamma=\frac{2}{3}$. We set $T^{off}=15$ and train with a batch size of $16$ and a learning rate of $1e^{-4}$. 

For online early prediction student training, we sample every $s$ frames from the entire sequences, with $s=3$. Each subsequence contains $\delta=5$ frames; as such, $s\delta=T^{off}=15$. We fix $T^{on}=\text{ceil}(\frac{N_{max}}{s\delta})$ for all sequences, with $N_{max}$ the maximum number of frames in a sequence across the entire dataset. Thus, every sequence can be studied in its entirety while training by batches. Shorter sequences are padded with black frames. Because the student loss (equation \ref{eq:student_loss}) is dynamic, the predictions for the black frames are not penalized. 

The R(2+1)D network is fine-tuned during training. As such, the 3D CNN backbone can shift its purpose from global to local feature extractor. The classifier, reused from the teacher, is frozen. We set hyperparameter $\eta=1$, batch size to $32$ and learning rate to $1e^{-3}$.

For OKDAD training, we keep $s$, $\delta$, $\eta$, batch size, learning rate, and frozen parameters identical as when training for online early prediction. We randomly sample subsequences of $T=40$ frame-blocks, which equals 20 seconds. The randomly sampled subsequences contain a mix of actions and non-actions. 

When evaluating the performances of OKDAD, predicted actionness with probabilities over 0.75 are considered positive. Temporal segments are created from adjacent positive predictions. Also, sequences are evaluated in their entirety to mimic a real-time scenario. An entire sequence from PKU-MMD averages around 3 minutes.

Previous works typically implement offline architectures for early prediction and action detection on NTU-RGB+D and PKU-MMD. To compare our online networks to these attempts, we calculate the offline classification accuracy by weighing the predictions at different time steps with the sigmoid weights from equation \ref{eq:student_loss}, with $T^{on}$ adapted to the observation ratio.

The Adam optimizer \cite{kingma2014adam} is used systematically across all training.

\subsection{NTU RBG+D human action recognition dataset}

NTU RGB+D is a state-of-the-art human action recognition dataset \cite{shahroudy2016ntu}. It contains 60 action classes, from daily activities to medical conditions, spread across 56,880 clips, 40 subjects, and 80 views. Each action contains up to two subjects. Captured from different views and setups, this great diversity makes NTU RGB+D a challenging dataset. Accuracy is used as the evaluation metric.

There are two benchmark evaluations: cross-subject (CS) and cross-view (CV). Following previous early prediction works on this dataset \cite{pang2019dbdnet}, \cite{wang2019progressive}, we only consider the CS benchmark, which splits training and test sets across different subjects. We sample 5\% of our training set as our validation set, as in \cite{shahroudy2016ntu}.

\setlength{\tabcolsep}{6pt}
\begin{table}[t]
\begin{center}
\caption{Action detection results on PKU-MMD (in mAP\textsubscript{a} \cite{liu2017pku}). Bold values are the best.}
\label{table:online_detection_results}
\begin{tabular}{c|cccc}
 & \multicolumn{4}{c}{ Cross-Subject } \\ 
\noalign{\smallskip}
\hline \hline
\noalign{\smallskip}
$\theta$ & 0.1 & 0.3 & 0.5 & 0.7 \\ 
\noalign{\smallskip}
\hline
\noalign{\smallskip}
RS+DR+DOF \cite{liu2017pku} & 0.647 & 0.476 & 0.199 & 0.026 \\
CS+DR+DOF \cite{liu2017pku} & 0.649 & 0.471 & 0.199 & 0.025 \\
TAP-B \cite{song2018spatio} & 0.544 & 0.514 & 0.461 & 0.327 \\
TAP-B-M \cite{song2018spatio} & 0.557 & 0.53 & 0.431 & 0.242 \\
RGB+D+F+S \cite{luo2018graph} & 0.903 & 0.895 & 0.833 & - \\ 
\noalign{\smallskip}
\hline
\noalign{\smallskip}
\textbf{OKDAD} $\alpha=1, \beta=0.5, \eta=0$ & 0.899 & 0.886 & 0.822 & 0.644 \\
\textbf{OKDAD} $\alpha=1, \beta=0.5, \eta=1$ & 0.908 & 0.893 & 0.826 & 0.651 \\
\textbf{OKDAD} $\alpha=0, \beta=0, \eta=0$ & 0.912 & 0.897 & 0.842 & 0.672 \\
\textbf{OKDAD} $\alpha=0, \beta=0, \eta=1$ & \textbf{0.915} & \textbf{0.905} & \textbf{0.850} & \textbf{0.679} \\ 

\end{tabular} 
\end{center}
\end{table}
\setlength{\tabcolsep}{1.4pt}

The results of our networks for different observation ratios are presented Table \ref{table:early_pred_results}. The strong performances of the teacher are outlined. For ratios greater than 20\%, the network consistently outperforms the previous state of the art by a large margin. At 50\%, accuracy is improved by 7.5\%, for a total of 76.0\%. At 100\%, the accuracy is improved by 4.9\%. On average, the teacher performs 2.5\% better than the best previous attempts. For very early predictions (less than or equal to 20\%), the teacher underperforms. At 10\%, we note a 17.4\% difference, and 6.3\% at 20\% compared to \cite{pang2019dbdnet}. 

Additionally, we report the offline performances of our online early prediction student. Best results are achieved with $\eta=1$ for the student with $\alpha=\beta=0$ for the teacher. The student is not able to match the performance of the teacher, except for an observation ratio of 100\% with 86.6\% for the student. Nonetheless, the student performs on par with the current state of the art, systematically outperforming it for ratios greater than 50\%. At 60\%, the student performs 1\% better than its closest competitor (with 75.5\%) and 5.1\% better at 100\% (with 86.6\%). These results are particularly encouraging considering the arguably harder online setting of our network.

\setlength{\tabcolsep}{4pt}
\begin{table*}[t]
\begin{center}
\caption{Impact of cosine penalty on teacher network. Accuracy in \% (Acc.), intraclass (Intra) and interclass (Inter) cosine similarity are reported. Bold values are the best.}
\label{table:teacher_ablation}
\begin{tabular}{c|c|c|c|c|c|c|c|c|c|c|c|c}
\multicolumn{2}{c}{ Observation ratio } & 10\% & 20\% & 30\% & 40\% & 50\% & 60\% & 70\% & 80\% & 90\% & 100\% & Avg. \\ 
\noalign{\smallskip}
\hline \hline
\noalign{\smallskip}
\multirow{3}{*}{ \shortstack{No penalty\\teacher} } & Acc. & 8.9 & \textbf{31.3} & \textbf{51.6} & 66.7 & 74.7 & 79.5 & 82.6 & 84.0 & 84.6 & 84.8 & 64.87 \\
& Intra & \textbf{0.80} & 0.49 & 0.44 & 0.46 & 0.50 & 0.53 & 0.55 & 0.56 & 0.57 & 0.57 & 0.55 \\
& Inter & 0.83 & 0.30 & 0.11 & 0.08 & 0.09 & 0.10 & 0.11 & 0.12 & 0.13 & 0.13 & 0.20 \\
\noalign{\smallskip}
\hline
\noalign{\smallskip}
\multirow{3}{*}{ \shortstack{Cos penalty\\teacher} } & Acc. & \textbf{10.5} & 29.5 & 49.9 & 66.7 & \textbf{76.0} & \textbf{81.1} & \textbf{83.9} & \textbf{85.4} & \textbf{86.3} & \textbf{86.4} & \textbf{65.57} \\ 
& Intra & 0.78 & \textbf{0.56} & \textbf{0.59} & \textbf{0.68} & \textbf{0.74} & \textbf{0.78} & \textbf{0.80} & \textbf{0.82} & \textbf{0.82} & \textbf{0.82} & \textbf{0.74} \\
& Inter & \textbf{0.73} & \textbf{0.25} & \textbf{0.10} & \textbf{0.07} & \textbf{0.06} & \textbf{0.06} & \textbf{0.06} & \textbf{0.06} & \textbf{0.06} & \textbf{0.06} & \textbf{0.15} \\
\noalign{\smallskip}
\end{tabular} 
\end{center}
\end{table*}
\setlength{\tabcolsep}{1.4pt}

\setlength{\tabcolsep}{4pt}
\begin{table*}[t]
\begin{center}
\caption{Contribution of layer reuse and knowledge distillation with different teachers. Accuracy in \% (Acc.), average cosine similarity (Sim.) and MSE between teacher and student feature vectors are reported. Bold values are the best.}
\label{table:student_ablation}
\begin{tabular}{cc|c|c|c|c|c|c|c|c|c|c|c|c}
\multicolumn{3}{c}{Observation ratio} & 10\% & 20\% & 30\% & 40\% & 50\% & 60\% & 70\% & 80\% & 90\% & 100\% & Avg. \\
\noalign{\smallskip}
\hline \hline
\noalign{\smallskip}
\multicolumn{2}{l|}{Baseline} & Acc. & 8.9 & 10.8 & 14.3 & 21.5 & 30.0 & 37.2 & 42.4 & 45.8 & 48.3 & 49.2 & 30.84 \\
\noalign{\smallskip}
\hline
\noalign{\smallskip}
\multirow{6}{*}{ \shortstack{Student with\\cos penalty\\teacher} } & \multirow{3}{*}{ $\eta=0$ } & Acc. & 18.5 & 26.3 & 37.4 & 53.9 & 67.0 & 73.9 & 78.5 & 81.7 & 83.4 & 84.8 & 60.54 \\
& & Sim. & 0.94 & 0.91 & 0.83 & 0.79 & 0.77 & 0.77 & 0.76 & 0.75 & 0.75 & 0.75 & 0.80 \\
& & MSE & 0.14 & 0.20 & 0.32 & 0.41 & 0.50 & 0.56 & 0.62 & 0.67 & 0.72 & 0.76 & 0.49 \\
\noalign{\smallskip}
\cline{2-14}
\noalign{\smallskip}
& \multirow{3}{*}{ $\eta=1$ } & Acc. & 18.0 & 27.4 & 36.9 & 52.3 & 65.8 & 73.8 & 79.1 & 82.5 & 84.4 & 85.5 & 60.57 \\
& & Sim. & 0.97 & 0.95 & 0.92 & 0.91 & 0.92 & \textbf{0.93} & \textbf{0.93} & \textbf{0.93} & \textbf{0.94} & \textbf{0.94} & \textbf{0.93} \\
& & MSE & \textbf{0.09} & \textbf{0.11} & \textbf{0.17} & \textbf{0.20} & \textbf{0.21} & \textbf{0.24} & \textbf{0.27} & \textbf{0.30} & \textbf{0.33} & \textbf{0.36} & \textbf{0.23} \\
\noalign{\smallskip}
\hline
\noalign{\smallskip}
\multirow{6}{*}{ \shortstack{Student with\\no penalty\\teacher} } & \multirow{3}{*}{ $\eta=0$ } & Acc. & 18.0 & 27.3 & 37.3 & 53.7 & 66.3 & 73.9 & 79.0 & 82.5 & 84.2 & 85.3 & 60.75 \\
& & Sim. & 0.94 & 0.90 & 0.80 & 0.74 & 0.70 & 0.69 & 0.68 & 0.68 & 0.67 & 0.67 & 0.75 \\
& & MSE & 0.63 & 0.71 & 0.87 & 0.81 & 0.83 & 0.85 & 0.86 & 0.89 & 0.91 & 0.93 & 0.83 \\
\noalign{\smallskip}
\cline{2-14}
\noalign{\smallskip}
& \multirow{3}{*}{ $\eta=1$ } & Acc. & \textbf{19.3} & \textbf{28.1} & \textbf{38.6} & \textbf{55.5} & \textbf{67.9} & \textbf{75.5} & \textbf{80.6} & \textbf{83.7} & \textbf{85.5} & \textbf{86.6} & \textbf{62.13} \\
& & Sim. & \textbf{0.98} & \textbf{0.97} & 0.92 & 0.89 & 0.88 & 0.87 & 0.87 & 0.87 & 0.87 & 0.86 & 0.90 \\
& & MSE & 0.48 & 0.51 & 0.54 & 0.43 & 0.40 & 0.40 & 0.41 & 0.42 & 0.44 & 0.46 & 0.45 \\
\noalign{\smallskip}
\end{tabular}
\end{center}
\end{table*}
\setlength{\tabcolsep}{1.4pt}

\subsection{PKU-MMD action detection dataset}

PKU-MMD is an RGB+D offline action detection dataset \cite{liu2017pku}. It contains 1,076 long sequences with approximately 20 actions from 51 classes per instance. It totals 21,545 actions performed by 66 subjects. Mean Average Precision of actions (mAP\textsubscript{a}) is used as the evaluation metric. To our knowledge, no online action detection dataset with infrared data exists. To compare our results with previous attempts, we evaluate the offline performances of OKDAD on the cross-subject benchmark, while evaluating the sequences in an online fashion.

Table \ref{table:online_detection_results} shows the performances of our OKDAD network for different temporal intersection over union thresholds $\theta$. Our method outperforms previous attempts for all thresholds. Especially for $\theta=0.7$, OKDAD does not experience a similar drop in performance as \cite{liu2017pku} and \cite{song2018spatio}. This suggests a considerable improvement in accurate temporal boundary detection, while still performing online. 

We evaluate OKDAD with two different teachers, with and without knowledge distillation for each. We find using a teacher without cosine similarity penalties yields better results. However, in both cases, encouraging cosine similarity between student and teacher vectors improves results noticeably. This is in line with our findings section \ref{sec:student_loss_ablation} (Knowledge distillation on online early prediction student).

Videos demonstrating the OKDAD network in action will be publicly available on the project page.

\subsection{Ablation studies}

We provide more experiments to better understand the different contributions of our knowledge distillation framework.

\subsubsection{Cosine similarity penalties on teacher learning}

During teacher training, intraclass similarity and interclass dissimilarity are encouraged via cosine penalty loss terms (\ref{eq:teacher_loss}). In Table \ref{table:teacher_ablation}, we compare the impact of the similarity loss terms on the test set. A "no penalty teacher" ($\alpha=\beta=0$) teacher is compared to its "cos penalty teacher" ($\alpha=1, \beta=0.5$) counterpart. Average intraclass and interclass cosine similarities are reported for different observation ratios. A value closer to 1 means similar orientation, closer to -1 means diametrically opposed. 

The loss terms from (\ref{eq:teacher_loss}) have the desired effects. When using the cosine terms, the intraclass similarity is considerably increased by 0.19 on average. The interclass similarity is adequately decreased by 0.05 on average. But it is interesting to note that the network pushes interclass vectors toward orthogonality without an explicit penalty. 

The cosine similarity loss terms also improve the accuracy scores, notably on observation ratios greater than 40\%. On average, accuracy is improved by 0.7\% and by about 1.5\% for ratios greater than 40\%. 

\subsubsection{Teacher layer reuse on online early prediction student}

Our distillation scheme consists of layer reuse (student with a teacher) and similarity learning ($\eta>0$). Table \ref{table:student_ablation} shows the impact of reusing teacher layers only (students with $\eta=0$) compared to our baseline. 

Our baseline is our early prediction student with a teacher trained on Kinetics-400 \cite{carreira2017quo} and without similarity learning ($\eta=0$). In other words, we use a generic feature extractor for our baseline, as done in previous works \cite{lin2017single} \cite{gao2017red}, \cite{wang2019progressive}, \cite{xu2019temporal}. Note that for the baseline, all 3D CNN parameters are frozen, as the network would not converge otherwise. This property is in favor of our method. Retraining the feature extractor for its new task before using it for the student allows it to be fine-tuned when training the student. 

For our students, we use two different teachers. The first one is trained with cosine similarity penalties, "cos penalty teacher" ($\alpha=1, \beta=0.5$), the second one without, "no penalty teacher". Note that the two teachers are identical to the ones used in Table \ref{table:teacher_ablation}. 

Compared to the baseline, average accuracy nearly doubles (30.84\% to about 60\% for both students). This clearly demonstrates the importance of an adequate feature extractor.

We also compare the cosine similarity between teacher and student feature vectors, as well as the MSE, at different observation ratios. It can be observed that a penalized teacher naturally increases similarity and decreases MSE between feature vectors. However, this does not seem to affect average performance as both student networks perform similarly. Interestingly, the student reusing the layers of the unpenalized teacher performs marginally better, especially in the later stages, even though the latter is less accurate as shown in Table \ref{table:teacher_ablation}. 

\subsubsection{Knowledge distillation on online early prediction student} \label{sec:student_loss_ablation}

Table \ref{table:student_ablation} also shows the contribution of enabling teacher guidance (students with $\eta>1$). 

As hypothesized, increasing cosine similarity between teacher and student feature vectors also reduces MSE without an explicit loss term. For the student with a cos penalty teacher, enabling teacher guidance increases similarity by 0.13 on average (0.80 to 0.93) and MSE reduces by half (0.49 to 0.23). Similar proportions can be observed using the unpenalized teacher. 

However, the increase in performance is marginal for the student with the cos penalty teacher (60.54\% to 60.57\% on average). More convincingly, the distillation scheme increases performances by 1.4\% on average (60.75\% to 62.13\%) using the no penalty teacher, more so for the later stages. Additionally, the student beats all teachers with that configuration for entire sequences, i.e. an observation ratio of 100\%. 

Indeed, the cos penalty teacher feature vectors are more convincingly approximated, but the tighter class clusters may also leave less room for error, explaining the lesser performances of the student in that case. 

\section{Conclusion}

Action recognition, early action recognition, and action detection are often studied separately. Moreover, it is important to distinguish between offline and online approaches. For the latter, temporal information, such as the duration of an action, are not used. In essence, online attempts mimic real-time scenarios. Previous early action prediction attempts are mostly offline. We argue that more efforts should be focused on online early prediction. Networks that have a strong ability to correctly classify an action as it happens should naturally perform well at detecting actions. In other words, online action detection can be decomposed into online early action recognition, to classify an ongoing action, and binary action detection, to detect if a current frame is part of an action or not.    

We propose a framework that builds upon this paradigm. Our online action detection network (OKDAD) is divided into a feature extractor in the form of a 3D CNN that analyses the current frames, the outputted feature vector is used to detect a potential ongoing action and classify it via parallel LSTM networks. The 3D CNN is first trained in the form of an offline early action prediction step and used as a teacher. It is then reused and fine-tuned as part of the end-to-end training of the OKDAD network. The OKDAD student benefits from knowledge distillation from teacher and student feature vectors cosine similarity.

Our method achieves state-of-the-art results on the NTU RBG+D and PKU-MMD datasets. This work focuses mainly on strong online early prediction performances. Our network achieves results on par with the current offline state-of-the-art attempts, despite performing online. To the best of our knowledge, we are among the first attempts to use knowledge distillation between networks performing different tasks, i.e. offline early prediction to online early prediction and online action detection. Also, our experiments are conducted on infrared data from RGB-D cameras which has never been attempted for early prediction and action detection.

\section*{Acknowledgment}

This work was supported by research funding from the Natural Sciences and Engineering Research Council of Canada, Prompt Québec, and industrial funding from Aerosystems International Inc. The authors would also like to thank their collaborators from Aerosystems International Inc.

\bibliography{bibliography}
\bibliographystyle{plain}

\newpage
\begin{IEEEbiography}[{\includegraphics[width=1in,height=1.25in,clip,keepaspectratio]{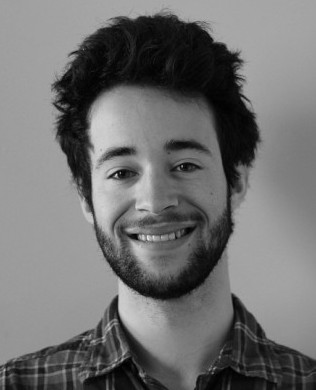}}]{Alban Main de Boissiere} is following jointly a M.Eng. at Institut National des Sciences Appliquées (INSA) Lyon and a M.A.Sc. at École de Technologie Supérieure (ETS) Montreal. His major interests include computer vision problems related to action recognition, early prediction, and online action detection. 
\end{IEEEbiography}

\begin{IEEEbiography}[{\includegraphics[width=1in,height=1.25in,clip,keepaspectratio]{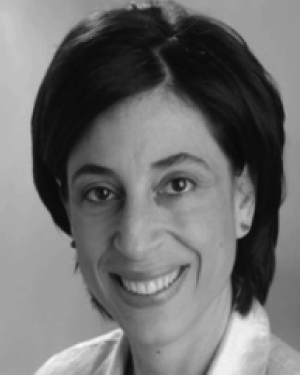}}]{Rita Noumeir} is a full professor at the Department of Electrical Engineering, at the École de Technologie Superieure (ETS) Montreal. Her main research interest is in applying artificial intelligence methods to create decision support systems. She has extensively worked in healthcare information technology and image processing. She has also provided consulting services in large-scale software architecture, healthcare interoperability, workflow analysis, technology assessment, and image processing, for several international software and medical companies including Canada Health Infoway.

Prof. Noumeir holds a Ph.D. and Master's degrees in Biomedical Engineering from École Polytechnique of Montreal.
\end{IEEEbiography}

\EOD

\end{document}